\documentclass[12pt]{article}
\usepackage{natbib}

\usepackage{amsmath}
\usepackage{amsfonts}
\usepackage{graphicx}
\usepackage{url} 
\usepackage{amsthm}
\usepackage{thmtools}
\usepackage{thm-restate}
\usepackage{algorithm2e}
\usepackage{algpseudocode}
\usepackage{subcaption}
\usepackage{comment}
\usepackage{enumitem} 
\usepackage{bm}
\usepackage{multicol} 
\usepackage{multirow} 
\usepackage{url}
\usepackage{xcolor}
\usepackage{listings} 
\definecolor{codegreen}{rgb}{0,0.6,0}
\definecolor{codegray}{rgb}{0.5,0.5,0.5}
\definecolor{codepurple}{rgb}{0.58,0,0.82}
\definecolor{backcolour}{rgb}{0.95,0.95,0.92}

\lstdefinestyle{mystyle}{
    backgroundcolor=\color{backcolour},   
    commentstyle=\color{codegreen},
    keywordstyle=\color{magenta},
    numberstyle=\tiny\color{codegray},
    stringstyle=\color{codepurple},
    basicstyle=\ttfamily\footnotesize,
    breakatwhitespace=false,         
    breaklines=true,                 
    captionpos=b,                    
    keepspaces=true,                 
    numbers=left,                    
    numbersep=5pt,                  
    showspaces=false,                
    showstringspaces=false,
    showtabs=false,                  
    tabsize=2
}

\usepackage{lstbayes}

\declaretheoremstyle[
  spaceabove=\topsep, spacebelow=\topsep,
  headfont=\normalfont\scshape,
  notefont=\mdseries, notebraces={(}{)},
  bodyfont=\normalfont,
  postheadspace=1em,
  qed=\qedsymbol
]{mythmstyle}
\declaretheorem[name=Proof, style=mythmstyle]{proofone}

\newcommand{\E}{{\rm I\kern-.3em E}}

\newcommand{\blind}{0}

\begin{document}

\SetKwInput{KwLoss}{Loss}
\SetKwInput{KwHp}{Hyperparameters}
\SetKwInput{KwParam}{Parameters}

\def\spacingset#1{\renewcommand{\baselinestretch}%
{#1}\small\normalsize} \spacingset{1}


\if0\blind
{
  \title{\bf Posterior Sampling of Probabilistic Word Embeddings}
  \author{Väinö Yrjänäinen\thanks{
    The authors gratefully acknowledge support from the Wallenberg AI, Autonomous Systems and Software Program (WASP) funded by the Knut and Alice Wallenberg Foundation. The computations were enabled by resources provided by the National Academic Infrastructure for Supercomputing in Sweden (NAISS), partially funded by the Swedish Research Council through grant agreement no. 2022-06725.}\hspace{.2cm}\\
    Department of Statistics, Uppsala University\\
    Isac Boström \hspace{.2cm}\\
    Department of Mathematical Sciences, \\ Chalmers University of Technology \\
    Måns Magnusson \hspace{.2cm}\\
    Department of Statistics, Uppsala University\\
    and \\
    Johan Jonasson \hspace{.2cm}\\
    Department of Mathematical Sciences, \\ Chalmers University of Technology}
  \maketitle
} \fi

\if1\blind
{
  \bigskip
  \bigskip
  \bigskip
  \begin{center}
    {\LARGE\bf Posterior Sampling of Probabilistic Word Embeddings}
\end{center}
  \medskip
} \fi

\bigskip
\begin{abstract}
Quantifying uncertainty in word embeddings is crucial for reliable inference from textual data. However, existing Bayesian methods such as Hamiltonian Monte Carlo (HMC) and mean-field variational inference (MFVI) are either computationally infeasible for large data or rely on restrictive assumptions.  
We propose a scalable Gibbs sampler using Polya-Gamma augmentation as well as Laplace approximation and compare them with MFVI and HMC for word embeddings. In addition, we address non-identifiability in word embeddings. Our Gibbs sampler and HMC correctly estimate uncertainties, while MFVI does not, and Laplace approximation only does so on large sample sizes, as expected. Applying the Gibbs sampler
to the US Congress and the Movielens datasets, we demonstrate the feasibility on larger real data. Finally, as a result of having draws from the full posterior, we show that the posterior mean of word embeddings improves over maximum a posteriori (MAP) estimates in terms of hold-out likelihood, especially for smaller sampling sizes, further strengthening the need for posterior sampling of word embeddings.
\end{abstract}

\noindent%
\vfill


\newpage
\spacingset{1.75} 

\section{INTRODUCTION}

Word embeddings are vector representations of words that capture statistical relationships within textual data. Traditional word embedding methods, such as 
word2vec
\citep{mikolov2013distributed} and GloVe \citep{pennington2014glove}, provide point estimates for each word, representing them as fixed vectors in a high-dimensional space. In recent years, word embeddings have been increasingly applied to inference tasks in the social sciences to investigate a wide range of linguistic and cultural aspects \citep{garg2018word, kozlowski2019geometry, rodman_2020, abramson2024inequality}.
However, standard word embedding methods (e.g., word2vec and GloVe) do not quantify the uncertainty associated with the embeddings, which is essential for reliable inference and interpretation in scientific settings. 


Probabilistic word embeddings \citep{rudolph2016exponential, bamler2017dynamic} extend traditional word embedding
and formulate word embeddings as a full probabilistic model. Nonetheless, quantifying uncertainty in these models poses significant challenges due to large datasets, non-identifiability, and large parameter spaces commonly involved in practical applications. An embedding model in typical settings can easily involve millions of parameters.

\citet{antoniak2018evaluating} proposed bootstrap for embedding uncertainty estimation. However, bootstrap estimation of uncertainty requires orders of magnitude more computation than obtaining point estimates. 
More recently, \citet{vallebueno2024statistical} proposed a method for quantifying uncertainty in the GloVe embeddings. This approach estimates the uncertainty conditional on the context embedding parameters and hence does not provide a rigorous approach to uncertainty estimation. In addition, the approach does not work for low-frequency words, which often constitute the bulk of words in natural language \citep{piantadosi2014zipf}.
In the Bayesian realm, \citet{bamler2017dynamic} proposed mean-field variational inference (MFVI) for the estimation of word embeddings. Although computationally efficient, it is well known that MFVI underestimates posterior variance \citep{wang2018frequentist}, limiting its use in scientific settings. On the other hand, Hamiltonian Monte Carlo (HMC) can be used for embeddings but is usually not feasible due to the high computational cost. Thus, while there has been work on embedding uncertainty, accurate Bayesian uncertainty quantification of embeddings is still elusive. 

In this paper, we present the following contributions:
\begin{enumerate}
    \item We study the identifiability of word embeddings in the context of Bayesian inference and prove that current embeddings are not identified. Furthermore, we present a conceptually easy and computationally feasible way to identify the skip-gram with negative samples (SGNS), one of the two popular word2vec models and prove that this approach identifies the model.
    \item We propose two new, computationally efficient algorithms for sampling embeddings: Laplace approximation and a Gibbs sampler. In addition, we compare with the gold-standard HMC on word embeddings. 
    \item We show that our proposed Gibbs sampler 
    is more accurate than MFVI, and more computationally efficient than HMC for large models and data, a common setting in social science applications. 
\end{enumerate}

\subsection{PROBABILISTIC WORD EMBEDDINGS} \label{sec:word_embeddings}

Probabilistic word embeddings \citep{rudolph2016exponential, bamler2017dynamic, rudolph2017dynamic} extend the standard SGNS word embedding model of \citep{mikolov2013distributed} to a probabilistic model. 

Let $\mathcal{D} = (w_i)_{i=1}^N$ denote a dataset consisting of a sequence of words, where each  word $w_i$ belongs to the \textit{vocabulary} $W = \{1, \dots, V\}$, the set of all relevant words in the corpus.\footnote{Throughout the article, we use an implicit mapping from the set of words $W = \{\texttt{word0}, \texttt{word1}, \dots \}$
onto indices $\{1, \dots, V\}$, e.g. $\{\texttt{dog} \mapsto 1, \texttt{cat} \mapsto 2, \texttt{cow} \mapsto 3\}$. Hence, the word vector $\rho_v$ for word $v \in W$ is just the $v$th column of the matrix $\rho$. The numeric indices have no meaning in word embeddings and thus no information is lost.}
For each word $w \in W$ we associate two embedding vectors: A \textit{target vector} $\rho_w \in \mathbb R^K$ and a \textit{context vector} $\alpha_w \in \mathbb R^K$, where $K$ denotes the dimensionality of the embedding. We organize the embeddings vectors into matrices such that $\rho \in \mathbb{R}^{V \times K}$ is the matrix of target vectors and $\alpha \in \mathbb{R}^{V \times K}$ the matrix of context vectors:
\begin{equation}
\rho = \begin{bmatrix}
           (\rho_{1})^\top \\
           (\rho_{2})^\top \\
           \vdots\\
           (\rho_{V})^\top
         \end{bmatrix}
,\qquad
\alpha = \begin{bmatrix}
           (\alpha_{1})^\top \\
           (\alpha_{2})^\top \\
           \vdots\\
           (\alpha_{V})^\top
         \end{bmatrix}.
\end{equation}
The complete set of parameters $\theta \in \mathbb{R}^{2V \times K}$ is then defined as
\begin{equation}
\theta = \begin{bmatrix}
           \rho \\
            \alpha 
         \end{bmatrix}.
\end{equation}

The SGNS likelihood nicely factors into terms, where each term only has a handful of parameters:
\begin{equation}
\label{eq:likelihood}
\begin{aligned}
\log p(\mathcal{D} \mid \theta) = 
\sum_{i=1}^N \biggl(
\underbrace{\sum_{v \in C^+_i} \log \sigma(\rho_{w_i}^\top\alpha_{v})}_{\text{positive samples}} 
+
&\underbrace{\sum_{v \in C^-_{i}} \log( 1 - \sigma(\rho_{w_i}^\top\alpha_{v}))}_{\text{negative samples}}
\biggr)
\end{aligned}
\end{equation}
where $C_i^+$ is the context window, or the set of positive samples, for word $w_i$ at index $i$ in the data. Specifically, the context window is the set of words within the distance $M \in \mathbb{N}$ from the center word, i.e. 
\begin{align}
C_i^+ = \{w_{i-M}, \dots, w_{i-1}, w_{i+1},\dots, w_{i+M}\}.
\end{align}

For each positive sample, $n_s \in \mathbb N$ negative samples forming $C_i^-$ are drawn randomly from the empirical distribution of all words in $W$, sometimes with slight adjustments \citep{mikolov2013distributed}. It can further be computationally and conceptually beneficial to formulate the likelihood as a sum over word pairs,
\begin{equation} \label{eq:ll_aggregated}
\begin{aligned}
\log p(\mathcal{D} \mid \theta) &= \sum_{(w, v) \in W \times W} \Big( n_{wv}^{+} \log \sigma( \rho_w^\top \alpha_v ) + n_{wv}^{-} \log( 1 - \sigma( \rho_w^\top \alpha_v ) ) \Big).
\end{aligned}
\end{equation}
Here, $n_{wv}^{+}$ and $n_{wv}^{-}$ denotes the number of times that $(w,v)$ appear in the positive and negative samples.

A common prior for the embedding vectors is a spherical Gaussian prior 
\begin{equation}
\begin{aligned}
    \rho_w &\sim \mathcal{N}(0, \lambda^{-1} \bm{I}), \text{ } w \in W,  \\
    \alpha_w &\sim \mathcal{N}(0, \lambda^{-1} \bm{I}), \text{ } w \in W \\
\end{aligned}
\end{equation}
where $\lambda \in \mathbb R_+$ is a hyperparameter. Other priors used for embedding vectors, and common in practical applications, include dynamic priors \citep{rudolph2017dynamic, bamler2017dynamic}, grouped priors \citep{rudolph2017structured}, and informative priors focused on interpretability \citep{bodell2019interpretable}, and graph priors \citep{yrjanainen2022probabilistic},  most of which are Gaussian. 

\section{IDENTIFICATION OF THE PROBABILISTIC SGNS MODEL} \label{sec:identifying}

The SGNS likelihood in Equation \ref{eq:likelihood} only consists of dot products between the target vectors $\rho_w$ and the context vectors $\alpha_v$. 
Because dot products are preserved under invertible linear transformations, we can apply the following  transformations for any invertible matrix $\bm{A}$ in the general linear group of degree $K$, denoted $\text{GL}(K)$, i.e., $\bm{A} \in \text{GL}(K)$:
\begin{equation}
    \rho^\prime_w = \bm{A}\rho_w, \quad
    \alpha^\prime_v = \bm{A}^{-\top}\alpha_v.
\end{equation}
Under these transformations the dot product is invariant, $ (\rho_w^\prime)^\top \alpha^\prime_v = (\rho_w)^\top \alpha_v $, and the likelihood remains unchanged. As a result, there are $K \times K$ degrees of freedom corresponding to these transformations that do not affect the likelihood. This makes the SGNS model non-identifiable in the sense that the mapping from the parameter space onto likelihoods is not one-to-one \citep{bickel2015mathematical}.

\citet{mu2019revisiting} propose adding $L_2$ regularization to identify the likelihood. However, their proof only shows that the MAP estimate is unique up to a rotation.
The likelihood still retains \textit{orthogonal} symmetries $p(\theta \mid x) = p(\theta \bm{O} \mid x)$, where $\bm{O}^\top\bm{O} =\bm{I}$.
Thus, Mu et al.'s \citeyearpar{mu2019revisiting} method does not properly identify the likelihood. 
Instead, this results in the posterior distribution exhibiting a \textit{spherical-shell} like structure, where even a very concentrated posterior will be distributed on the shell of a hypersphere, see Figure \ref{fig:sim_donuts}. Intuitively one might view this as any orientation of the embeddings is valid.
 
Fixing $K^2$ elements of $\alpha$
eliminates these symmetries. We consider $\theta$ for which the $K$ first context vectors, $\alpha_I$, where $I = \{1, \dots K\}$, form a fixed and predetermined invertible matrix $\bm{M} \in \mathbb R^{K \times K}$, while the rest of the context vectors form an unrestricted matrix $\alpha_J \in \mathbb R^{(V-K) \times K}$, $J = \{K +1, \dots, V\}$. In Proposition \ref{proposition:surjective}, we show that the restricted model can represent any likelihood that the unrestricted model can, and in Proposition \ref{proposition:injective} we show that the likelihood of the restricted model is identifiable.

\begin{restatable}{prop}{propositionsurjective} \label{proposition:surjective}
Given any embedding $\hat \alpha, \hat \rho$, and any invertible matrix $\bm{M} \in \mathbb{R}^{K \times K}$, the following embedding
\begin{equation}
    \begin{aligned}
    \rho = \hat \rho (\hat \alpha_{I}^{-1} \bm{M})^{-\top}, \quad \alpha = \begin{bmatrix} \bm{M} \\ \hat \alpha_J \hat \alpha_{I}^{-1} \bm{M} \\ \end{bmatrix} \\
    \end{aligned}
\end{equation}
has the same likelihood
\begin{equation}
    p(\mathcal{D} \mid \rho, \alpha ) = p(\mathcal{D} \mid \hat \rho, \hat \alpha ) 
\end{equation}
for any dataset $\mathcal{D}$.
\end{restatable}
Proof in Appendix \ref{appendix:proofs}.

\begin{restatable}{prop}{propositioninjective}  \label{proposition:injective}
Given any invertible matrix $\bm{M} \in \mathbb{R}^{K \times K}$, any two embeddings $\alpha = \begin{bmatrix} \bm{M} \\ \alpha_J \\ \end{bmatrix}, \rho$ and $\alpha^\prime = \begin{bmatrix} \bm{M} \\ \alpha_J^\prime\end{bmatrix}, \rho^\prime$, for which $\alpha \neq \alpha^\prime$ or $\rho \neq \rho^\prime$, the likelihood
\begin{equation}
    p(\mathcal{D} \mid \rho, \alpha ) \neq p(\mathcal{D} \mid \hat \rho, \hat \alpha ) 
\end{equation}
for some dataset $\mathcal{D}$.
\end{restatable}
Proof in Appendix \ref{appendix:proofs}.

Without loss of generality, we can choose any set of $K$ context vectors (or word vectors) to fix to have the same result. The identifiability correction allows for direct comparisons between parameter values. In the Bayesian realm, this allows us to calculate for example $\hat{R}$ values to assess convergence to the stationary distribution and effective sample sizes for specific parameters. It also makes it possible to calculate a posterior mean $\theta$, which would otherwise just be zero, as $p(\theta \mid \mathcal{D}) = p(-\theta\mid \mathcal{D})$ for any $\theta$.


\subsection{COMPARING IDENTIFIED AND NON-IDENTIFIED EMBEDDINGS}


Rotation-invariant features are commonly used to analyze word embeddings. These include Euclidean distances between the words (column distances) and cosine similarity \citep{garg2018word, eisenstein2018natural, kozlowski2019geometry}.
The co-occurrence probability of any word pair $(w,v)\in W\times W$ is calculated as 
\begin{equation} \label{eq:co_occ_one_pair}
    P(w \land v \mid \theta) = \sigma(\alpha_w^\top \rho_v).
\end{equation}
Using this, we define the distance between two embeddings as the root mean squared error (RMSE) of the co-occurrence probabilities over all word pairs
\begin{equation} \label{eq:co-occ-divergence}
\begin{aligned}
    \text{RMSE}_{co}(\theta, \theta^\prime) = \sqrt{\frac{1}{V^2}\sum_{v,w \in W}(P(w \land v \mid \theta) -  P(w \land v \mid \theta^\prime))^2}
\end{aligned}
\end{equation}
Notably $\text{RMSE}_{co}$ is invariant under invertible transformations $\bm{A}$. When $\theta = \theta^\prime$ then $\text{RMSE}_{co}(\theta, \theta^\prime) = 0$, but when $\text{RMSE}_{co}(\theta, \theta^\prime) = 0$, $\theta^\prime = \begin{bmatrix}\rho \bm{A} \\ \alpha \bm{A}^{-1}\end{bmatrix}$ for some transformation $\bm{A} \in \text{GL}(K)$. Hence, we can compare unidentified as well as identified embeddings. We go on to use it as a similarity metric for two embeddings. This is also the basis of our convergence analysis when true parameters are present. We interpret $\text{RMSE}_{co}(\hat \theta, \theta_{true}) = 0$ as the estimate $\hat \theta$ having converged to the true parameter $\theta_{true}$. 

\section{POSTERIOR APPROXIMATION METHODS}

We propose two approximation methods for probabilistic word embeddings. Additionally, we show how Laplace approximation can be utilized efficiently for a subset of the vocabulary.

\subsection{LAPLACE APPROXIMATION} \label{sec:laplace}

Laplace approximation of the posterior
where it is difficult to sample directly. In Laplace approximation, we approximate the posterior with a multivariate Gaussian distribution around the MAP, i.e.
\begin{equation}
\theta \sim  \mathcal{N}(\hat{\theta}_\textit{MAP}, - \bm{H}^{-1}(\hat{\theta}_\textit{MAP}))\,,
\end{equation}
where $\bm{H}(\theta)$ is the Hessian matrix of the posterior at $\theta$. To estimate distances between words, we need the covariances of all entries in these word vectors. The most straightforward way to do this would be to calculate the inverse of the Hessian of the posterior at $\hat{\theta}_\text{MAP}$. For instance, to calculate the credible interval of the distance between $\rho_\texttt{dog}$ and $\rho_\texttt{cat}$ under the Laplace approximation, only the covariance matrix of $[\rho_\texttt{dog}^\top, \rho_\texttt{cat}^\top]$ is needed.
Due to the large parameter space, it is often not possible to invert $\bm{H}$ in reasonable time. 



Under the word embedding model, it is possible to exploit the sparsity and structure of the Hessian to make Laplace approximation more scalable due to the sparsity of the likelihood and the prior. The structure of the sparse Hessian is further elaborated on in Appendix \ref{appendix:hessian}.

\subsection{GIBBS SAMPLING}  \label{sec:gibbs}

For the SGNS likelihood, the distribution of $\rho_w, w \in W$ is, conditionally on $\alpha$, independent of the other $\rho_{w^\prime}, w^\prime \in W, w^\prime \neq w$. This holds similarly for $\alpha_w$ as well
\begin{equation}
\begin{aligned}
    p(\mathcal{D} \mid \rho, \alpha) &= \prod_{w \in W} p(\mathcal{D} \mid \rho_w, \alpha) = \prod_{v\in W} p(\mathcal{D} \mid \rho, \alpha_v)
\end{aligned}
\end{equation}
Moreover, conditionally on $\alpha$ (or similarly on $\rho$), the likelihood for $\rho$ is identical to logistic regression when treating $\alpha$ as covariates,
\begin{equation}
p(\mathcal{D} \mid \rho_w, \alpha) \propto \prod_{v \in W} \sigma(\alpha_v^\top \rho_w)^{n^+_{wv}} \sigma(-\alpha_v^\top \rho_w)^{n^-_{wv}}.
\end{equation}


Since the distribution of each $\rho_v$ for any $v \in W$ is analogous to the posterior in logistic regression, it is straightforward to construct a Gibbs sampler. The Gibbs sampler alternates sampling $\rho$ and $\alpha$, where each $\rho_v$, and each $\alpha_v$ is sampled independently from each other, lending itself well to parallel computation, hence scaling to real-word data sizes. Logistic regression parameters with Gaussian priors can be directly sampled via Polya-Gamma latent variables \citep{polson2013bayesian}. Together, this results in a simple and parallel blocked Gibbs sampler as described in Algorithm \ref{alg:gibbs}.

\begin{algorithm}[h!]
\caption{Gibbs sampling of word embeddings (SGNS) }
\label{alg:gibbs}
\KwData{$x \in \mathcal X$}
\KwHp{Dimensionality $K \in \mathbb N$, vocabulary $W$, \\vocabulary size $V = \lvert W \rvert$, number of iterations $N \in \mathbb N$, number of Polya-Gamma iterations $S \in \mathbb N$}
\KwParam{$\rho, \alpha \in \mathbb R^{V \times K}$}
\KwLoss{$\mathcal L : \mathcal X \times \mathbb R^{2V \times K} \to \mathbb R = \sum_{w \in W} \log f_w(\rho_w, \alpha) = \sum_{w \in W} \log g_w(\alpha_w, \rho)$ (SGNS)}
\For{$w \in W$}{
$\rho_{w,0} \gets \mathcal{N}(0, 1/K)$ \\
$\alpha_{w,0} \gets \mathcal{N}(0, 1/K)$
}
\For{$n \in \{1, \dots, N\}$}{
\For{$w \in W$}{
    Calculate conditional prior $\mathcal{N}(\mu_{\rho_w}, \Sigma_{\rho_w})$ for $\rho_w$ given $\rho_{-w}$ and $\alpha$ \\
    $\alpha^\prime \gets (\alpha_{v_j}, \dots, \alpha_{v_J})$, where $j$ are the indices where $w_j = w$ \\
    $x^\prime \gets (x_{j}, \dots, x_{J})$ \\
    $\beta_w \gets \mathcal{N}(0, 1/K)$ \\
    \For{$s \in \{1, \dots, S\}$}{
    Sample $\omega \sim \text{PolyaGamma}(n_i, \alpha^{\prime \top} \beta_w)$ \\
    Sample $\beta_w\sim N(\mu_\omega, V_\omega)$, where $V_\omega = \left(\alpha^{\prime \top} \text{diag}(\omega)\alpha^\prime+ \Sigma_{\rho_w}^{-1}\right)^{-1}$ and $\mu_\omega = V_\omega(\alpha^{\prime \top}(x_w - 1/2 ) + \Sigma_{\rho_w}^{-1}\mu_{\rho_w})$\\
    }
    $\rho_{w,n} \gets \beta_w$ \\
    }\For{$w \in W$}{
    Sample $\alpha_{w,n}$ with a similar procedure as $\rho_{w_n}$ \\
    }
}
\end{algorithm}

\section{EXPERIMENTS}
We study the Laplace approximation in Section \ref{sec:laplace} and the Gibbs sampler in Section \ref{sec:gibbs}, and compare them with current state-of-the-art for estimation, namely MAP estimation, MFVI, and HMC sampling.
We use HMC implemented in Stan\footnote{\url{https://mc-stan.org/}} and the MFVI algorithm presented by \citet{bamler2017dynamic}, using the Stan implementation of MFVI (see details in Appendix \ref{appendix:mfvi_and_hmc_implementation}). All code is provided on GitHub\footnote{\url{https://anonymous.4open.science/r/embedding-uncertainty-estimation-096B/README.md}}.

To study convergence and uncertainty estimates, we use a simulated dataset with known parameters. We complement the simulation experiments with empirical experiments with real data, namely two standard datasets used for embedding methods, the MovieLens dataset \citep{harper2015movielens} and the US Congress dataset \citep{gentzkow2018congressional}.
 
\clearpage 
\subsection{SIMULATED DATA}

In the simulation study, we use a set of randomly generated word and context vectors. For each $w \in W$, the word and context vectors are simulated from
\begin{equation}
\begin{aligned}
    \rho_w \sim \mathcal N(0, \frac{\varepsilon^2}{K}\bm{I} ), \quad
    \alpha_w \sim \mathcal N(0, \frac{\varepsilon^2}{K} \bm{I}), \\
\end{aligned}
\end{equation}
where $K$ is the dimensionality of the embedding, and $\varepsilon=1$ the signal-to-noise ratio.
The embedding is scaled inversely to the dimensionality so that the dot products, and thus the signal-to-noise ratio, remain similar in magnitude regardless of the dimensionality $K$.

We generate random word pairs $(w_i, v_i) \in W \times W$ by sampling uniformly. A Bernoulli random variable, representing whether the pair is a positive or negative sample, $X_{i}$ is then sampled for $i \in \{1, \dots, N\}$
\begin{equation}
\begin{aligned}
    (w_i, v_i) &\sim \text{Uniform}(W \times W) \\
    X_i &\sim \text{Bernoulli}(\sigma(\alpha_{v_i}^\top \rho_{w_i}))
\end{aligned}
\end{equation}
Each resulting dataset is then estimated with the true $K$ and $\lambda$, using MAP estimation, VI,  HMC and Gibbs sampler.

In natural language, words often have frequencies proportional to their inverse rank in the vocabulary, i.e. words tend to follow Zipf's law \citep{zipf1935psychobiology, piantadosi2014zipf}. As another experiment, we simulate this by generating $w_i$ and $v_i$ with 
\begin{equation}
\begin{aligned}
    p(w_i) &\propto \frac{1}{\text{rank}(w_i)^a + b}\,, \quad
    p(v_i) &\propto \frac{1}{\text{rank}(v_i)^a + b}\,,\\
\end{aligned}
\end{equation}
where $\text{rank}( \cdot )$ is an enumeration of $W$, and $a$ and $b$ are positive constants. We set $a=1$ and $b=2.7$, as suggested by \citet{mandelbrot1953informational}. 

\subsubsection{Elimination of Rotational Invariances}

\noindent
As discussed in Section \ref{sec:identifying}, the likelihood can be identified by fixing a submatrix of $\alpha$. However, any set of $K$ context vectors can be selected, and their values can be set to any set of linearly independent vectors.
In practice, we first find a MAP estimate $\hat \theta_\textit{MAP}$, then we set $K$ context vectors to their values during sampling. We found that the selection of words had no effect on the results (see Appendix \ref{appendix:additional_plots}), so we used the \textit{last} $K$ context vectors.

\begin{figure}[h]
\begin{center}
     \includegraphics[width=0.4\columnwidth]
     {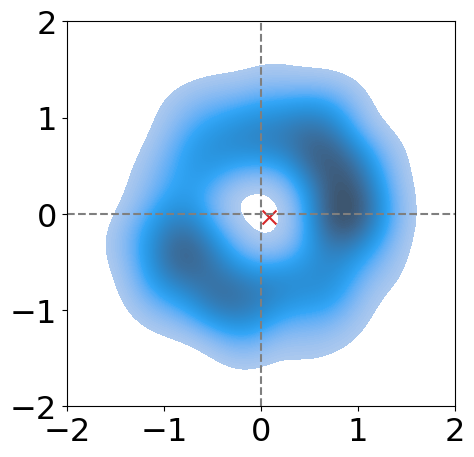}
     \qquad
     \includegraphics[width=0.4\columnwidth]
     {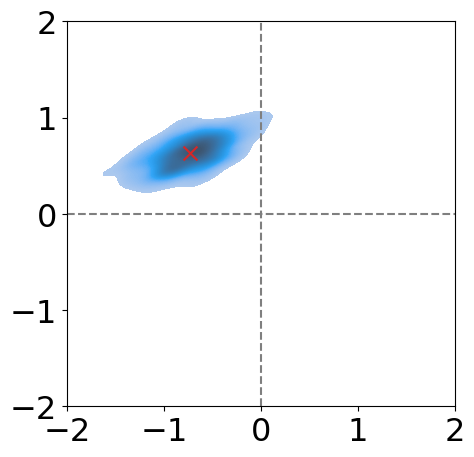}
    \caption{With $K=2$, the marginal distribution of $\rho_{1,1}$ and $\rho_{1,2}$ displays a circular shell-like symmetry (left), which is eliminated by fixing $\alpha_I$ to a MAP estimate (right). The posterior expectation is marked with a red cross. $N=2 \cdot 10^4$, $V=10$. Samples drawn using Stan HMC.
    }
    \label{fig:sim_donuts}
\end{center}
\end{figure}

In Figure \ref{fig:sim_donuts}, we illustrate how the non-identifiability of the posterior distributions results in rotational symmetries, and then show the effect of identifying the posterior.
The marginal posterior of $\rho_{1,1}$ and $\rho_{1,2}$ shows the theorized spherical-shell posterior geometry.
The distribution is symmetric around the origin, and has a zero mean. Further we show how restricting $K \times K$ context vectors successfully eliminates the symmetries, simplifying the geometry and making the posterior mean nonzero. Further illustrations of the spherical shell for higher dimensions are provided in Appendix \ref{appendix:additional_plots}.

\subsubsection{Convergence, Coverage and Effective Sample Size}

\begin{figure}[!h]
    \centering
    \includegraphics[width=0.46\linewidth]
    {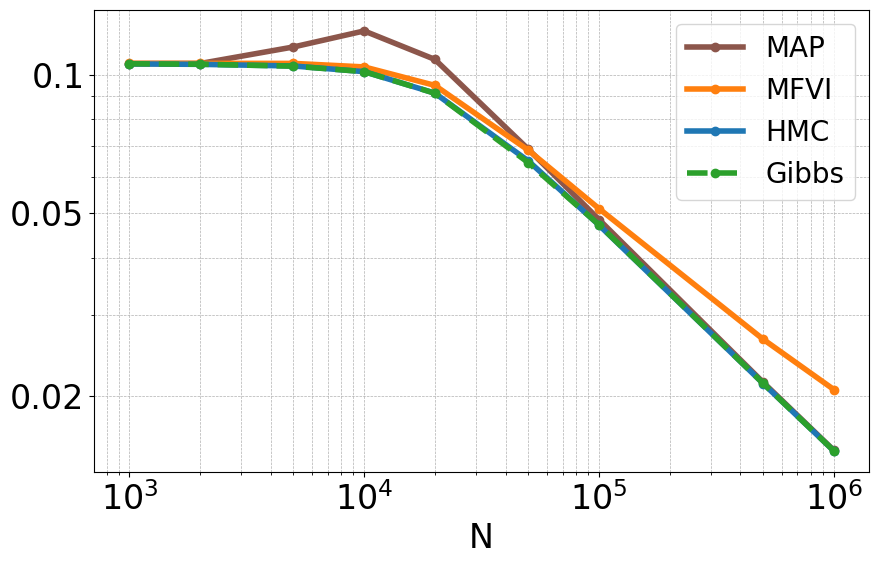}
    \quad
    \includegraphics[width=0.46\linewidth]{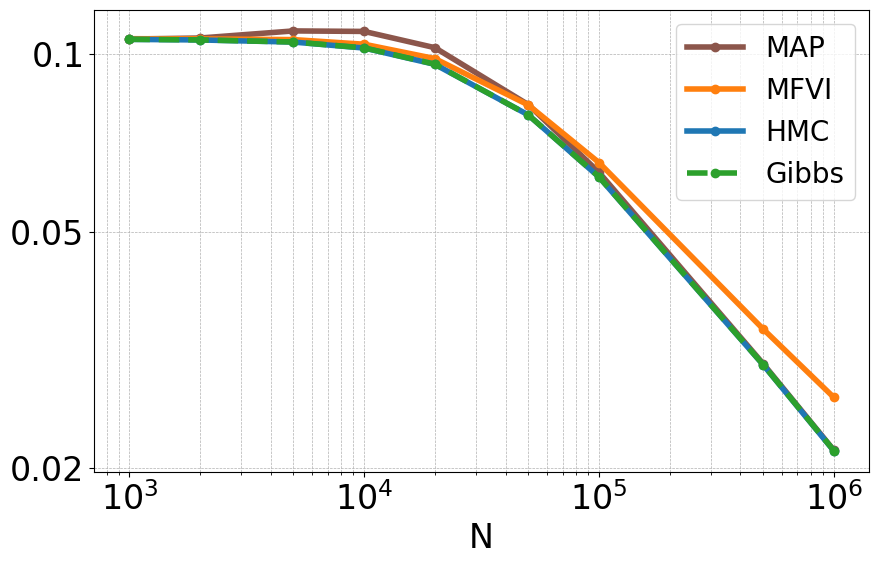}
    \caption{Log-log plotted $\text{RMSE}_{co}$ for $K=5$ and $V=100$. Averaged over 10 simulated datasets. The left plot shows the results for the uniform simulation the right plot shows the results for the Zipf's law simulation.
    }
    \label{fig:sim_K5_V100_rmse_loglog}
\end{figure}

For the simulated data, we study the convergence of the estimated embeddings $\hat \theta$ to the true value $\theta_{true}$. To do this, we use the divergence $\text{RMSE}_{co}$ defined in Equation (\ref{eq:co-occ-divergence}). Figure \ref{fig:sim_K5_V100_rmse_loglog} shows the convergence of the different estimation methods in terms of $\text{RMSE}_{co}$. HMC and the Gibbs sampler perform similarly, while MFVI appears to converge slightly slower as $N$ increases. With small sample sizes, MAP underperforms the other methods, while on large sample sizes it approaches the performance of HMC and the Gibbs sampler. Asymptotically, as expected, HMC, MAP and the Gibbs sampler are very close.

\begin{figure}[h]
    \centering
    \includegraphics[width=0.46\linewidth]{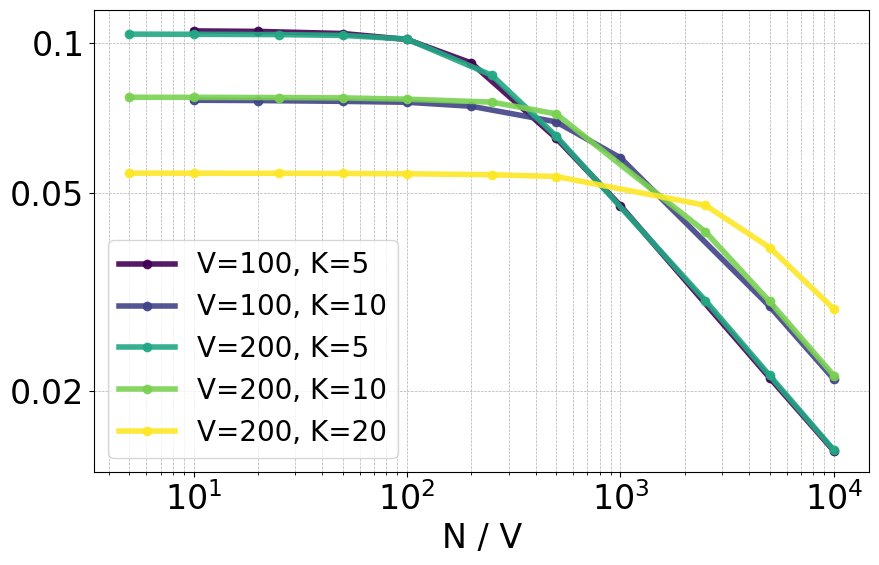}
    \caption{Log-log plotted $\text{RMSE}_{co}$ by $N/V$ for the different data sizes, using the Gibbs sampler. Averaged over 10 simulated datasets, uniform simulation.}
    \label{fig:sim_rmse_loglog}
\end{figure}

As seen in Figures \ref{fig:sim_K5_V100_rmse_loglog} and \ref{fig:sim_rmse_loglog}, a loglinear pattern in $\text{RMSE}_{co}$ emerges once $N$ is sufficiently large. This is consistent across different values of $V$ and $K$. Log-log linearity corresponds to $O (N^{-C})$ convergence of $\text{RMSE}_{co}$ in terms of $N$ for some positive constant $C$.
For all combinations of $K$ and $V$, numerical estimates of $C$ were close to $-0.5$, i.e. $\text{RMSE}_{co}(N) \approx O(N^{-0.5})$. Since Bayesian point estimates $\hat \beta$ approach a $1/\sqrt{N}$ scaled Gaussian as $N \to \infty$ \citep{kleijn2012bernstein}, the squared error $\mathbb{E}[(\hat \beta -\beta)^2]$ is $O(N^{-1})$. Thus its square root is $O(N^{-0.5})$ which matches the convergence rate that we observe for HMC, MAP, and the Gibbs sampler, but interestingly not for MFVI. Moreover, the increase in $\text{RMSE}_{co}$ given an increase in $V$ was very close to 1, implying a convergence rate of $O(N^{0.5}/V)$. This also holds for the Zipf-simulated data.

\begin{table*}[h!]
\centering
\small
\caption{Coverage ($\%$) of the true co-occurrence probability $P(w \land v\mid \theta_\textit{true})$ of the $90\%$-credible interval of $P(w \land v\mid \theta)$ in the simulated experiment for different data sizes. Averaged over 10 simulated datasets and all word pairs $(w, v) \in W \times W$. 
}
\scalebox{0.98}{
\begin{tabular}{|c|l|llllllllll|}
\hline
Dataset   & Method & 1k   & 2k  & 5k  & 10k & 20k  & 50k   & 100k & 500k & 1M &\\
\hline
\multirow{3}{*}{$K=5$, $V=100$} & HMC & 89.9 & 89.8 & 89.7 & 89.7 & 89.7 & 89.9 & 89.9 & 90.2 & 89.9 &\\
& MFVI  & 90.2 & 88.3 & 84.0 & 84.5 & 84.2 & 85.8 & 86.6 & 83.3 & 79.5 &\\
& Gibbs & 89.7 & 89.6 & 89.7 & 89.5 & 89.6 & 89.9 & 89.6 & 90.1 & 89.7 &\\ 
& Laplace  & 94.9 & 94.6 & 98.6 & 97.7 & 96.4 & 94.9 & 91.9 & 90.5 & 90.3 &\\
\hline
\hline
\multirow{3}{*}{$K=10$, $V=100$} 
& HMC   & 90.0 & 90.0 & 90.0 & 90.0 & 90.0 & 90.1 & 90.2 & 90.0 & 90.1  &\\
& MFVI  & 92.5 & 91.9 & 90.4 & 88.0 & 83.7 & 85.4 & 85.6 & 85.1 & 83.0 &\\
& Gibbs & 89.9 & 89.9 & 89.9 & 89.9 & 89.9 & 89.9 & 89.6 & 89.9 & 90.1 &\\
& Laplace  & 89.9 & 89.9 & 89.7 & 89.4 & 98.0 & 96.5 & 95.2 & 90.8 & 90.1 &\\
\hline
\hline
\multirow{3}{*}{$K=20$, $V=100$} 
& HMC   & 90.0 & 89.9 & 90.0 & 89.9 & 90.0 & 90.0 & 89.9 & 89.7 & 89.9 &\\
& MFVI  &93.3 & 93.2 & 92.8 & 92.3 & 90.9 & 87.3 & 83.5 & 85.3 & 84.5 &\\
& Gibbs  & 89.4 & 89.4 & 89.4 & 89.4 & 89.4 & 89.4 & 89.4 & 89.7 & 90.0 &\\     
& Laplace   & 79.9 & 79.9 & 79.6 & 79.4 & 33.6 & 64.9 & 95.7 & 91.1 & 89.4  &\\
\hline
\hline
\multirow{3}{*}{$K=5$, $V=200$} 
& HMC  & 90.2 & 90.2 & 90.2 & 90.3 & 90.2 & 90.2 & 90.1 & 90.1 & 90.0 & \\
& MFVI & 91.5 & 90.6 & 87.7 & 84.0 & 84.5 & 83.9 & 86.2 & 85.4 & 83.1  & \\
& Gibbs & 90.1 & 90.1 & 90.1 & 90.1 & 90.0 & 90.0 & 89.9 & 89.7 & 89.9  &\\                    
& Laplace  & 97.5 & 97.5 & 97.8 & 99.5 & 98.7 & 97.9 & 96.3 & 91.2 & 90.5 &\\
\hline
\hline
\multirow{3}{*}{$K=10$, $V=200$} 
& HMC  & 88.7 & 88.8 & 88.8 & 88.8 & 88.9 & 89.0 & 89.3 & 90.4 & 90.2 & \\ 
& MFVI  & 92.3 & 92.1 & 91.4 & 90.0 & 87.4 & 82.6 & 84.4 & 85.6 & 84.9 &\\
& Gibbs &89.5 & 89.5 & 89.5 & 89.6 & 89.7 & 89.7 & 89.7 & 89.7 & 89.7 & \\
& Laplace   & 95.0 & 95.0 & 95.0 & 94.6 & 94.8 & 99.1 & 97.9 & 94.1 & 91.7 &\\
\hline
\hline
\multirow{3}{*}{$K=20$, $V=200$} 
& HMC   & 90.1 & 90.1 & 90.2 & 90.1 & 90.2 & 90.2 & 90.1 & 90.1 & 90.1 & \\                 
& MFVI  & 93.5 & 93.5 & 93.5 & 93.2 & 92.6 & 90.5 & 87.7 & 85.1 & 84.4  &\\
& Gibbs & 90.1 & 90.1 & 90.1 & 90.1 & 90.1 & 90.1 & 90.0 & 89.8 & 89.8 & \\         
& Laplace & 89.8 & 90.0 & 90.0 & 89.8 & 89.8 & 89.4 & 94.8 & 96.5 & 94.9 &\\

\hline
\end{tabular}
}
\label{tab:coverage}
\end{table*}

According to Bernstein-von-Mises theorem, under relatively mild assumptions, credible intervals should have valid coverage proportions as the number of observations $N \to \infty$ \citep{kleijn2012bernstein}. Hence, we study the coverage of $P(w \land v \mid \theta)$ as a function of data size. The results over all word pairs and 10 simulated datasets are shown in Table \ref{tab:coverage}.
HMC and the Gibbs sampler cover the true value roughly 90\% of the time regardless of $K, V$ or $N$. Laplace approximation yields incorrect confidence intervals for small $N$. This may be both due to an inaccurate normal approximation, and to numerical errors in the inversion and decomposition of the Hessian, something we observed during our experiments. For large values of $N$, the Laplace approximation confidence intervals approach accurate coverage of 90\%, which is expected based on the Bernstein-von-Mises theorem \citep{kleijn2012bernstein}. For MFVI, the confidence intervals are too wide for small values of $N$, and for large values of $N$, MFVI only covers the true value under 80.0\%-85.0\% of the time. 
This issue worsens as $N$ grows and follows a similar pattern across all combinations of $K$ and $V$. The tendency of MFVI to underestimate uncertainty is in line with theoretical results on variational inference \citep{wang2018frequentist}. 

\begin{table}[h] 
\centering
\small
\caption{Median effective sample size, out of 1000 draws after a burn-in of 1000. Simulated data, uniform co-occurrences, $K=5, V=100$. }
\begin{tabular}{|llllll|}
\hline
$N$ & 1K & 2K & 5K & 20K & 50K  \\
\hline
HMC  & 886.6 & 913.8 & 931.6 & 964.3 & 993.7  \\
Gibbs  & 579.8  & 327.9  & 207.0  & 286.9  & 547.7 \\
\hline
\end{tabular}
\label{tab:ess}
\end{table}

In Table \ref{tab:ess} we compare the effective sample size (ESS), as defined via the autocorrelation \citep{gelman2013bayesian}, for HMC and Gibbs. As expected, HMC yields higher effective sample sizes, reaching 89\% - 99\% of the nominal sample size, while the Gibbs sampler achieves 20-60\%, depending on the dataset size.
The sampling rates are shown in Table \ref{tab:throughput_sim}. While HMC is faster for small models, the Gibbs sampler outperforms HMC on the two largest simulation settings. As we will see, running HMC on large datasets is practically not possible due to poor scaling with respect to model and data size. 

\begin{table}[h]
\centering
\small
\caption{Samples per second for one simulated dataset with uniform co-occurrences. Gibbs results for $S=10$. Words per second is samples per seconds normalized by $V$.
}
\begin{tabular}{|l|ll|ll|}
\hline
Method & $N$ & $K \times V$ & Samples/s & Words/s\\
\hline
HMC & $5 \cdot 10^4$ & $5 \times 100$ & 17.700 & 1769.9 \\
Gibbs& $5 \cdot 10^4$  & $5 \times 100$ & 3.333  & 333.3  \\
\hline
HMC & $5 \cdot 10^4$ & $10 \times 200$ & 4.762 & 942.4 \\
Gibbs & $5 \cdot 10^4$ & $10 \times 200$ & 2.467  & 493.3 \\
\hline
HMC & $5 \cdot 10^4$ & $20 \times 200$ & 2.802 & 560.4 \\
Gibbs & $5 \cdot 10^4$ & $20 \times 200$ & 2.233 & 446.7 \\
\hline
HMC & $5 \cdot 10^5$ & $5 \times 100$ & 2.951 & 295.1  \\
Gibbs& $5 \cdot 10^5$  & $5 \times 100$ & 2.900 & 290.0   \\
\hline
HMC & $5 \cdot 10^5$ & $10 \times 200$ & 2.386 & 478.0  \\
Gibbs & $5 \cdot 10^5$ & $10 \times 200$ & 2.880 & 576.0  \\
\hline
HMC & $5 \cdot 10^5$ & $20 \times 200$ & 0.506 & 101.2 \\
Gibbs & $5 \cdot 10^5$ & $20 \times 200$ & 0.733 & 146.7  \\
\hline
\end{tabular}
\label{tab:throughput_sim}
\end{table}

\begin{figure}[!h]
    \centering
    \includegraphics[width=0.46\linewidth]{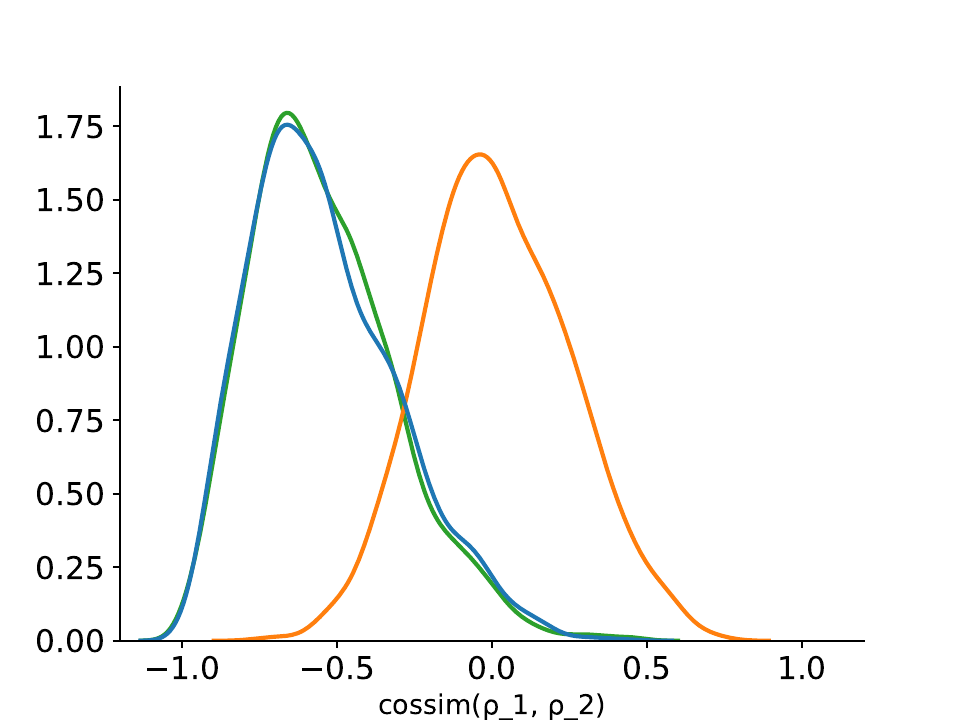} 
    \includegraphics[width=0.46\linewidth]{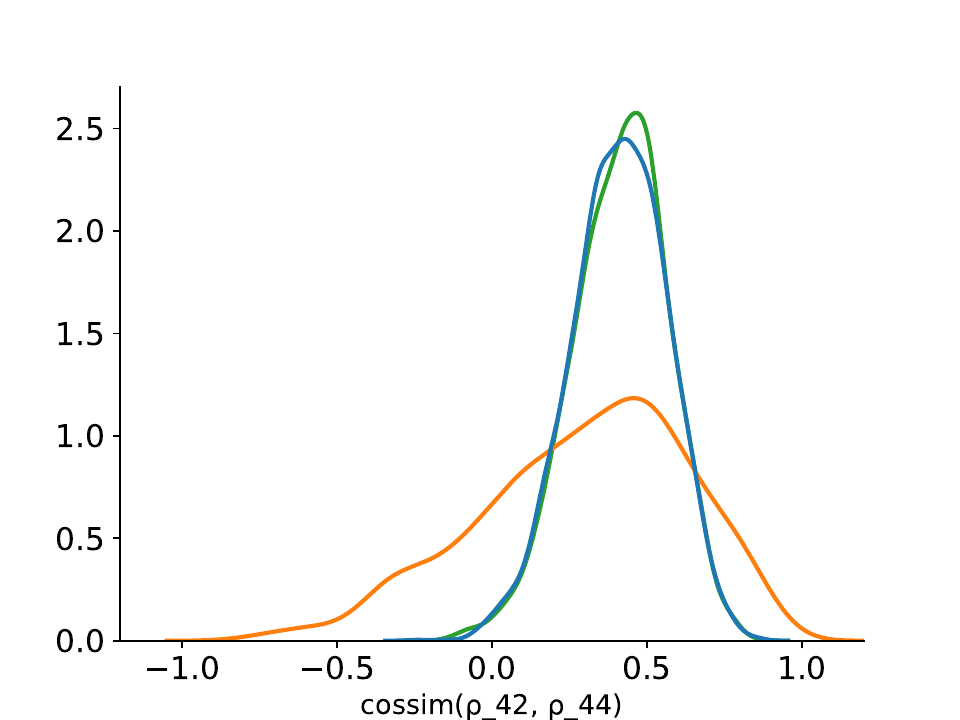}
    \caption{Above: Cosine similarity between $\rho_1$ and $\rho_2$ in the first simulated dataset. Below: Cosine similarity between $\rho_{42}$ and $\rho_{44}$. $K=5, V=100, N=10^{5}$. Gibbs sampler in green, HMC in blue and MFVI in orange.}
    \label{fig:sim_K5_V100_cossim}
\end{figure}

Finally, we compare the posterior distributions of cosine similarities, a common use case for word embedding in applied settings \citep{garg2018word, eisenstein2018natural, kozlowski2019geometry}.
As seen in Figure \ref{fig:sim_K5_V100_cossim}, the posterior distributions of the cosine similarities obtained from HMC and the Gibbs sampler are very similar. However, the posteriors obtained from MFVI differ from these drastically.
This is not surprising, due to the use of MFVI posterior approximation, and clearly shows the problem of using MFVI for uncertainty estimation in SGNS embeddings.

\subsection{MOVIELENS DATASET}

\citet{rudolph2016exponential} embed the MovieLens data \citep{harper2015movielens} using a probabilistic embedding to find thematically similar movies. Following this, we use the MovieLens data as a basis for non-text embedding model experiments. The data contains a collection of numerical movie reviews, which were preprocessed to be suitable for SGNS likelihood in a similarly to \citet{rudolph2016exponential}. \clearpage

\begin{table}[h!]
\caption{Samples per second for the Movielens dataset. Above $K=20$, $V=603$, $N=5 \cdot 10^{4}$, and below $K=100, V=603, N=10^{6}$. Words per second is samples per second normalized by $V$. Gibbs results for $S=10$. Runs with an asterisk (*) did not finish before a time limit.}
\centering
\small
\subfloat[][$K=20$]{
\begin{tabular}{|l|l|ll|}
\hline
Method & $K$ & Samples/s & Words/s  \\
\hline
HMC & 20 & 0.8333 & 502.5 \\
Gibbs & 20 & 0.278  & 167.5 \\
\hline
\end{tabular}}
\qquad
\subfloat[][$K=100$]{
\begin{tabular}{|l|l|ll|}
\hline
Method & $K$ & Samples/s & Words/s  \\
\hline
HMC & 100 & $< 0.0069$* & $< 4.19$* \\
Gibbs & 100 & 0.0267  & 16.15 \\
\hline
\end{tabular}}
\label{tab:throughput_movielens}
\end{table}


MovieLens is a relatively small dataset, allowing us to compare the performance of HMC and the Gibbs sampler. For smaller embedding models ($K=20$), HMC is still competitive, as shown in Table \ref{tab:throughput_movielens}. However, HMC's efficiency drastically drops as dimensionalities and dataset size increase.
HMC only produced a handful of samples before the program was terminated due to hitting the time limit.
While the Gibbs sampler also yields significantly less samples per second in these conditions, its performance is still acceptable. It should however be noted that Gibbs sampling benefits from parallel updates, whereas the standard Stan HMC implementation does not.


\begin{table}[h]
\centering
\caption{Averaged hold-out log likelihood on the MovieLens data by estimator and sample size. $K=20, \lambda=1$.}
\begin{tabular}{|l|lll|}
\hline
 & $\hat \theta_{\textit{PM-Gibbs}}$ & $\hat \theta_{\textit{MAP}}$ & $\hat \theta_{\textit{MFVI}}$  \\
\hline
$N=85,000$ & \bf{-0.7320} & -0.7801 & -1.1142 \\
$N=170,000$ & \bf{-0.6814} & -0.7247  & -1.1201 \\
$N=425,000$ & \bf{-0.6458} & -0.6721  & -0.8715 \\
$N=850,000$ & \bf{-0.6382} & -0.6571  & -0.6609 \\
\hline
\end{tabular}\label{tab:holdout_movielens}
\end{table}

Another benefit of using posterior sampling is that we can use the posterior mean as a point estimates of the word embeddings instead of MAP.
In Table \ref{tab:holdout_movielens}, we compare the posterior mean estimator $\hat \theta_{\textit{PM}}$ to the MAP estimator $\hat \theta_{\textit{MAP}}$ and the MFVI mean $\hat \theta_{\textit{MFVI}}$ on the MovieLens data. The metric used is the SGNS likelihood on held-out data. On all sample sizes ($N$), the posterior mean estimator outperforms the MAP estimator.

\subsection{US CONGRESS DATASET}

For the text data experiment, we use the US Congress speech corpus of \citet{gentzkow2018congressional}, a well-known text dataset popular in the social sciences \citep{myrick2021external, weiss2021political, card2022computational}. We restrict the vocabulary to the 5000 most common words in the dataset, and use a subset of $1-10$ million words. Further hyperparameters were the window size $M=2$, and the number of negative samples $n_s=1$. After this preprocessing, we obtained a sample of 10 million observations. We set $K=100$ and $\lambda = 1$, common value in applications.

Our experiments are twofold. First, we compare the Gibbs sampler to HMC and MFVI in terms of posteriors of cosine similarities. Furthermore, we compare the VI mean, MAP and HMC and Gibbs sampler posterior mean point estimates in terms of test set hold-out performance. For the larger data sizes, we do not use HMC as it was infeasibly slow. 

\begin{figure}[h]
\begin{center}
    \includegraphics[width=0.46\columnwidth]{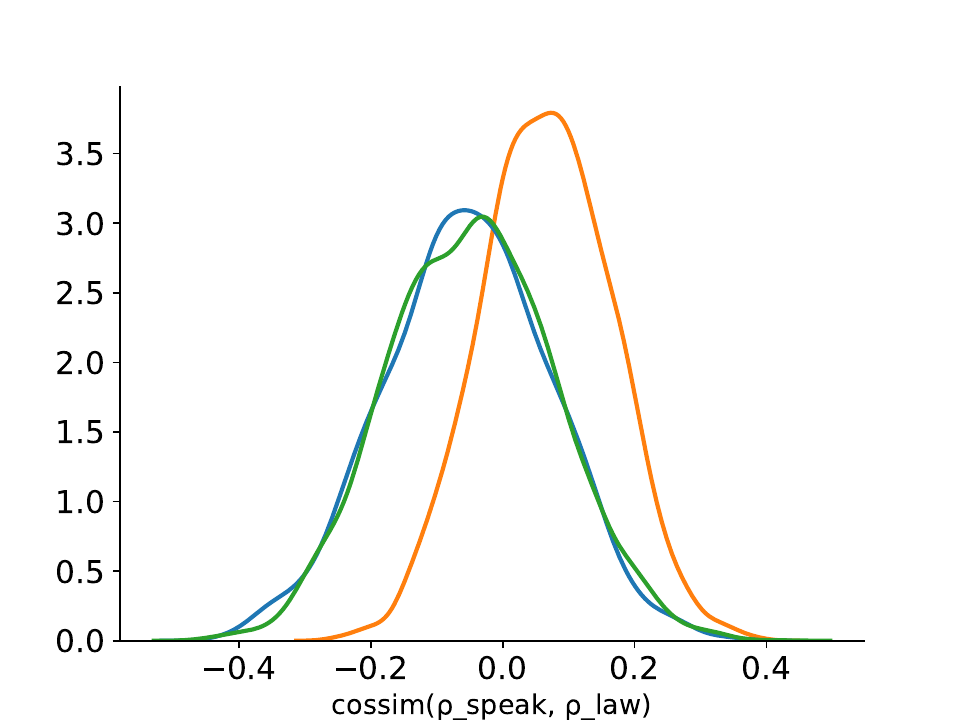} 
    \includegraphics[width=0.46\columnwidth]{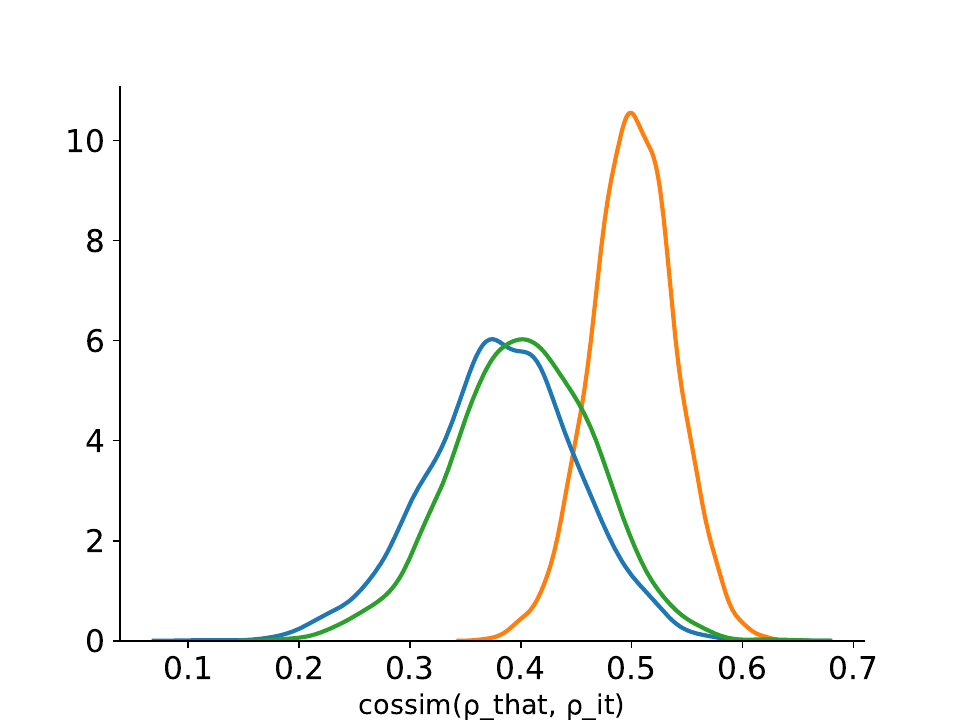} \\
    \caption{Posterior distribution of the cosine similarity between words for the Gibbs sampler (green), HMC (blue) and MFVI (orange). Left: 'speak' and 'law', right: 'that' and 'it'. $K=50, N=500,000$.}
    \label{fig:cossim_speak_law_k50}
\end{center}
\end{figure}

\begin{figure}[h]
\begin{center}
    \includegraphics[width=0.46\columnwidth]{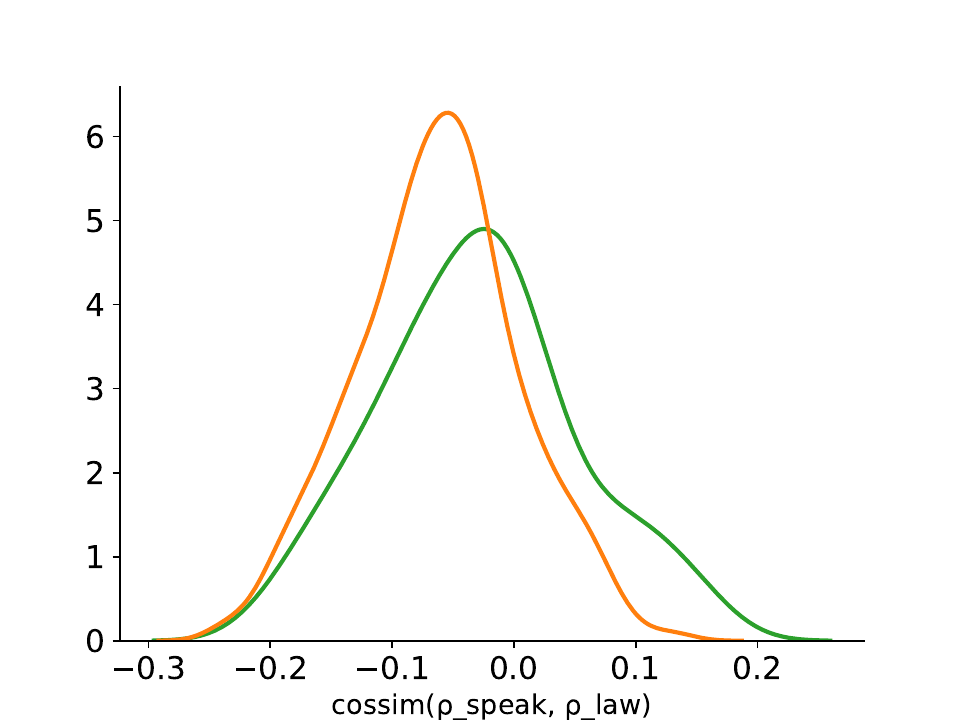} 
    \includegraphics[width=0.46\columnwidth]{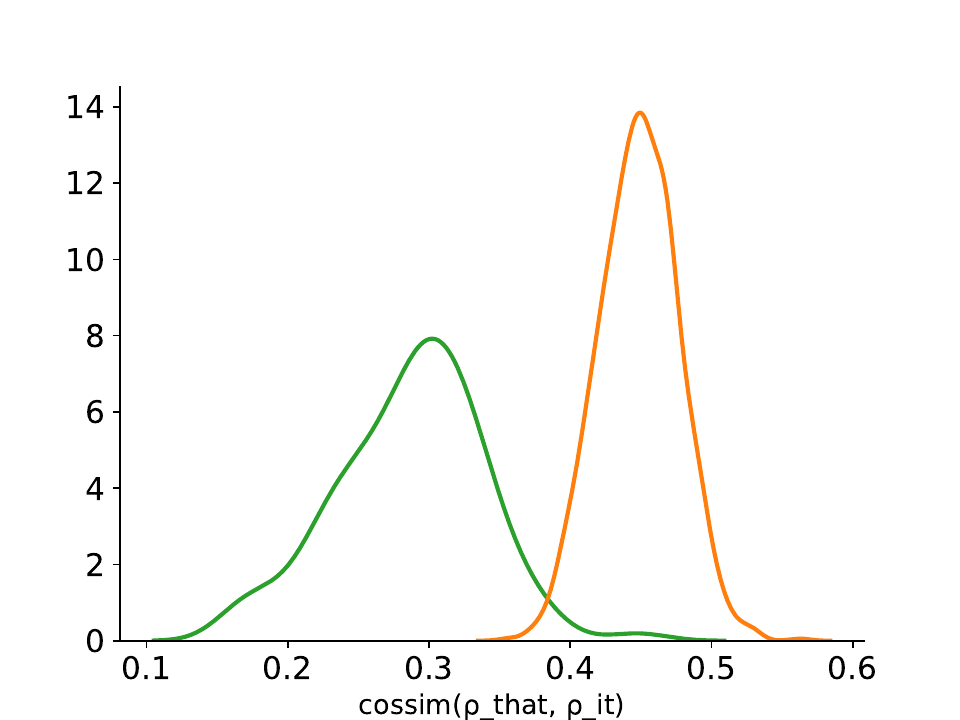} \\
    \caption{Posterior distribution of the cosine similarity between words for the Gibbs sampler (green) and MFVI (orange). Left: 'speak' and 'law', right: 'that' and 'it'. $K=100, N=500,000$.}
    \label{fig:cossim_speak_law}
\end{center}
\end{figure}

The posteriors of the cosine similarities between some word pairs are illustrated in Figure \ref{fig:cossim_speak_law_k50}, on a low dimensionality $K=50$ where all three algorithms could be run. The Gibbs sampler and HMC yield similar posteriors for this metric. On the other hand, MFVI yields substantially different results from HMC and the Gibbs sampler, both in terms of the postirior mean, and  a smaller posterior variance.
The latter result is in line with results on MFVI \citep{wang2018frequentist}.
As seen in Figure \ref{fig:cossim_speak_law}, the results are similar when $K=100$.
The posteriors are narrower for MFVI than for the Gibbs sampler, and they also differ in means.
Since cosine similarity is one of the primary ways of analysis of word embeddings, these differences argue against the use of mean-field approximation for SGNS embeddings.

\begin{table}[htb!]
\centering
\caption{Hold-out log likelihood on the US Congress data. $K=50$ on the left, $K=100$ on the right. $\lambda=1, M=2, n_s=1$. PM-Gibbs is the posterior mean estimator for the Gibbs sampler, PM-HMC is the posterior mean estimator for the HMC sampler.}
\subfloat[][$K=50$]{\begin{tabular}{|l|lll|}
\hline
 & $\hat \theta_{\textit{PM-Gibbs}}$  & $\hat \theta_{\textit{PM-HMC}}$ & $\hat \theta_{\textit{MAP}}$ \\
\hline
$N=10^{5}$ & -0.7648 & \bf{-0.7584}  & -0.8927  \\
$N=5 \cdot 10^5$ & \bf{-0.7556} & -0.7574 & -0.8648 \\
$N=10^6$ & \bf{-0.72549} & -0.7374 & -0.7896 \\
$N=5 \cdot 10^6$ & \bf{-0.6551} & NA & -0.6625 \\
\hline
\end{tabular}}
\qquad
\subfloat[][$K=100$]{\begin{tabular}{|l|ll|}
\hline
 & $\hat \theta_{\textit{PM-Gibbs}}$ & $\hat \theta_{\textit{MAP}}$ \\
\hline
$N=5 \cdot 10^5 $ & \textbf{-0.7374} & -0.8209   \\
$N=10^6$ & \textbf{-0.7136} & -0.7836  \\
$N=5 \cdot 10^6$ & \bf{-0.6594} & -0.6948  \\
$N=10^7$ & -0.6360 & \textbf{-0.6357}\\
\hline
\end{tabular}}
\label{tab:holdout_us}
\end{table}

Finally, Table \ref{tab:holdout_us} presents the results of the hold-out performance experiment on the US congress data. On both $K=50$ and $K=100$, both posterior mean estimators
outperform the MAP estimator. However, on the full training data $N=10^7$, MAP is, as expected, on par with the posterior mean point estimator. On $K=50$, where we can compare the posterior mean estimators from HMC and the Gibbs sampler, they perform similarly.

\section{CONCLUSION}
In this paper, we showed that the undesirable symmetries in SGNS embeddings persist despite the previous efforts to eliminate them. We then proposed a simple way to eliminate the symmetries and identify the SGNS model. 
This allows for full use of $\hat{R}$ and the effective sample size, and makes the posterior mean a reasonable point estimate.

On simulated data, we showed that the current standard method for estimating probabilistic SGNS, MFVI, yields inaccurate credible intervals, especially on large data. We also showed that HMC and the Gibbs sampler yield accurate credible intervals, and Laplace approximation yields accurate credible intervals on larger sample sizes.

On real data, we demonstrated that MFVI and the Gibbs sampler yield different posterior distributions of cosine similarities, which are one of the primary metrics used in word embeddings. 
This suggests that MFVI is not a good approximation for SGNS embeddings
in terms of uncertainty estimates.
We also showed that the Gibbs sampler is computationally feasible in this setting, while  Stan HMC is not. Finally, we showed that the new posterior mean estimator outperformed the MAP and VI mean estimators.

\if0\blind
{
} \fi

\if1\blind
{
} \fi

\clearpage

\section*{Data Availability Statement}

The data that support the findings of this study are openly available in the \textit{MovieLens} data repository at \url{https://grouplens.org/datasets/movielens/}, as well as the 
\textit{Congressional Record for the 43rd-114th Congresses: Parsed Speeches and Phrase Counts} data at
\url{https://data.stanford.edu/congress_text}.

\bibliographystyle{chicago}
\bibliography{refs}



\newpage

\onecolumn

\title{Posterior Sampling of Probabilistic Word Embeddings \\(Supplementary Material)}
\maketitle

\appendix

\section{Proofs} \label{appendix:proofs}

\propositionsurjective*

\begin{proofone}
For any data $\mathcal{D}$, the likelihood is a function of the product $\alpha \rho^\top$
\begin{equation}
    p(\mathcal{D} \mid \rho, \alpha) \equiv g(\alpha \rho^\top)
\end{equation}
which in turn means that
\begin{equation}
\alpha \rho^\top = \alpha^\prime (\rho^\prime)^\top \implies p(\mathcal{D} \mid \rho, \alpha) = p(\mathcal{D} \mid \rho^\prime, \alpha^\prime)
\end{equation}
for any data $\mathcal{D}$. Thus, we need to show that the matrices $\alpha \rho^\top$ and $\hat \alpha \hat \rho^\top$ are always equal
\begin{align}
\alpha \rho^\top &= \begin{bmatrix} \bm{M} \\ \hat \alpha_J \hat \alpha_{I}^{-1} \bm{M} \\ \end{bmatrix} ( \hat \rho (\hat \alpha_{I}^{-1} \bm{M})^{-\top})^\top = \begin{bmatrix} \bm{M} \\ \hat \alpha_J \hat \alpha_{I}^{-1} \bm{M} \\ \end{bmatrix} ( \hat \alpha_{I}^{-1} \bm{M})^{-1}\hat \rho \\
&= \begin{bmatrix} \bm{M} \\ \hat \alpha_J \hat \alpha_{I}^{-1} \bm{M} \\ \end{bmatrix} \bm{M}^{-1} \hat \alpha_{I} \hat \rho^\top = \begin{bmatrix} \bm{M}\bm{M}^{-1} \hat \alpha_{I} \\ \hat \alpha_J \hat \alpha_{I}^{-1} \bm{M} \bm{M}^{-1} \hat \alpha_{I}\\ \end{bmatrix} \hat \rho^\top = \begin{bmatrix} \hat \alpha_{I} \\ \hat \alpha_J \\ \end{bmatrix} \hat \rho^\top = \hat \alpha \hat \rho^\top
\end{align}
and thus $p(\mathcal{D} \mid \hat \rho, \hat \alpha) \equiv p(\mathcal{D} \mid \rho, \alpha)$.
\end{proofone}

\propositioninjective*

\begin{proofone}
\begin{equation}
    \theta \neq \theta^\prime \implies p(\mathcal{D} \mid \theta) \neq p(\mathcal{D} \mid \theta^\prime)
\end{equation}
By contraposition, this is equivalent to
\begin{equation}
    p(\mathcal{D} \mid \theta^\prime) = p(\mathcal{D} \mid \theta) \implies \theta = \theta^\prime 
\end{equation}
The likelihood is a function of the product
\begin{equation}
    p(\mathcal{D} \mid \theta) \equiv g(\alpha \rho^\top)
\end{equation}
which means that two likelihoods are identical for any data $\mathcal{D}$ iff $\alpha \rho^\top = \hat \alpha \hat \rho^\top$. Given that $\alpha_I = \hat \alpha_I = \bm{M}$, the equality becomes
\begin{align}
\alpha \rho^\top &= \begin{pmatrix}
\alpha_I \rho_I^\top & \alpha_I \rho_J^\top\\
\alpha_1 \rho_I^\top & \alpha_1 \rho_J^\top
\end{pmatrix} = \begin{pmatrix}
\hat \alpha_I \rho_I^\top & \hat \alpha_I \rho_J^\top\\
\alpha_1 \rho_I^\top & \alpha_1 \rho_J^\top
\end{pmatrix}  = \begin{pmatrix}
\hat \alpha_I \hat \rho_I^\top & \hat  \alpha_I \hat \rho_J^\top\\
\hat \alpha_1 \hat \rho_I^\top & \hat \alpha_1 \hat \rho_J^\top
\end{pmatrix}
\end{align}
Since $\alpha_I^{-1}$ exists
\begin{align}
\hat \alpha_I \rho_I^\top = \hat \alpha_I \hat \rho_I^\top \implies \rho_I = \hat \rho_I \\
\hat \alpha_I \rho_J^\top = \hat \alpha_I \hat \rho_J^\top \implies \rho_J = \hat \rho_J
\end{align}
and thus
\begin{align}
\alpha_J \hat \rho_I^\top = \hat \alpha_J \hat \rho_I^\top  \implies \alpha_J = \hat \alpha_J
\end{align}
i.e., $\rho = \hat \rho$ and $\alpha = \hat \alpha$.
\end{proofone}

\section{Structure of the Hessian } \label{appendix:hessian}

\begin{figure}[h]
\begin{center}
\includegraphics[width=0.4\linewidth]{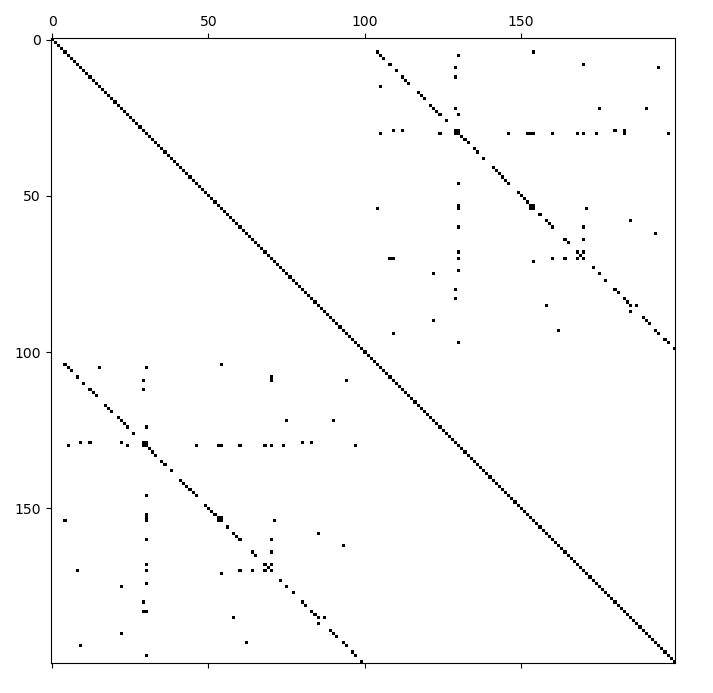}
\end{center}
\caption{Nonzero elements of the Hessian matrix for a subset of 100 word and 100 context vectors in a model. Data from 50M token Wikipedia subset, standard preprocessing by \citet{mikolov2013distributed} was used.}
\centering
\label{fig:sparsity}
\end{figure}

As seen in Equation \eqref{eq:likelihood}, the skip-gram likelihood factors into terms with only a few parameters. Generally, in natural text corpora, most of the possible word pairs do not appear together. Consequently, the interactions between words are inherently sparse. This means that most of the entries in the Hessian are going to be zero (for the positive samples). For example, in a random 50 million token subset of Wikipedia, roughly 0.25\% of the word pairs actually co-occur within a window size of 5. The sparsity is illustrated in Figure \ref{fig:sparsity}. We can leverage this property to improve computations of the Hessian.

The Hessian of the positive samples also factors into submatrices consisting of outer products
\begin{equation}
    \frac{\partial^2}{\partial^2 \rho_w}\log p(\mathcal{D} \mid \theta)^+ = - \sum_{v \in W} \alpha_v \alpha_v^\top \sigma^\prime(\rho_w^\top \alpha_v )  n^+_{v,w}
\end{equation}
for the diagonal blocks, and for the off-diagonal blocks
\begin{align}
    \frac{\partial^2}{\partial \rho_w \partial \alpha_v}\log p(\mathcal{D} \mid \theta)^+ &= 
     - \rho_w \alpha_v^\top \sigma^\prime(\rho_w^\top \alpha_v) n_{w,v}^+ - I \sigma(\rho_w^\top \alpha_v)n_{w,v}^+ \\
     \frac{\partial^2}{\partial \rho_w \partial \rho_v}\log p(\mathcal{D} \mid \theta)^+ &= \mathbf{0} \\
     \frac{\partial^2}{\partial \alpha_w \partial \alpha_v}\log p(\mathcal{D} \mid \theta)^+ &= \mathbf{0}
\end{align}
where the positive samples $n_{w,v}^+$ 
will often be 0. Since the derivative of the logistic function $\sigma^\prime(\cdot)$ is a symmetric function, the same formulas hold for the negative samples as well.
Note that the diagonal submatrices are low-rank if there are less than $K$ co-occuring word types, and the off-diagonal submatrices are always rank 1 or 0.



\section{Additional Results} \label{appendix:additional_plots}

\begin{figure}[h!]
    \centering
    \includegraphics[width=0.4\linewidth]{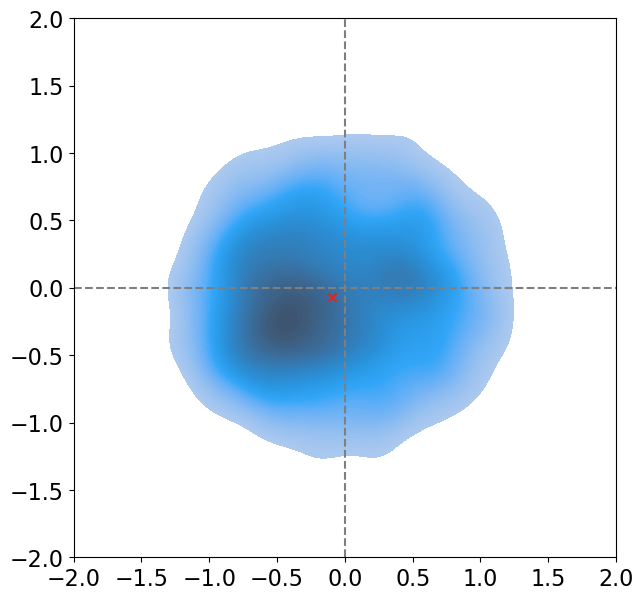}
    \includegraphics[width=0.4\linewidth]{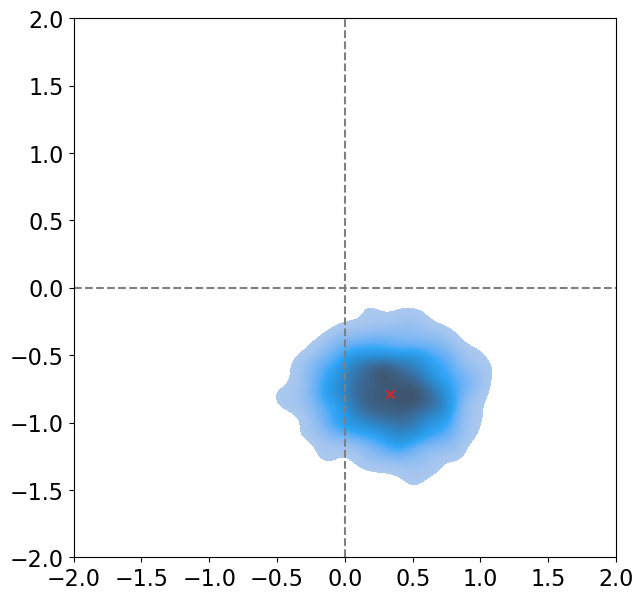}\\
    \caption{$\rho_{4,1}$ and $\rho_{4,2}$ for $K=5$, $V=100$, $N=50,000$. The left figure demonstrates the rotational symmetry which is eliminated by fixing $\alpha_I$ to a MAP estimate, shown in the right figure.
    }
    \label{fig:symmetry_K5}
\end{figure}
In higher dimensions than $K=2$, the hypersphere resulting from the orthogonal symmetries $\bm{O}$ is not a circle, but a sphere or a hypersphere instead. Hence, the 2D projection of the circular shell is closer to standard Gaussian than a torus, as in the $K=2$ case. The symmetries for $K=5$ are illustrated in Figure \ref{fig:symmetry_K5}. 

\begin{figure}[h]
\begin{center}
    \includegraphics[width=0.46\columnwidth]{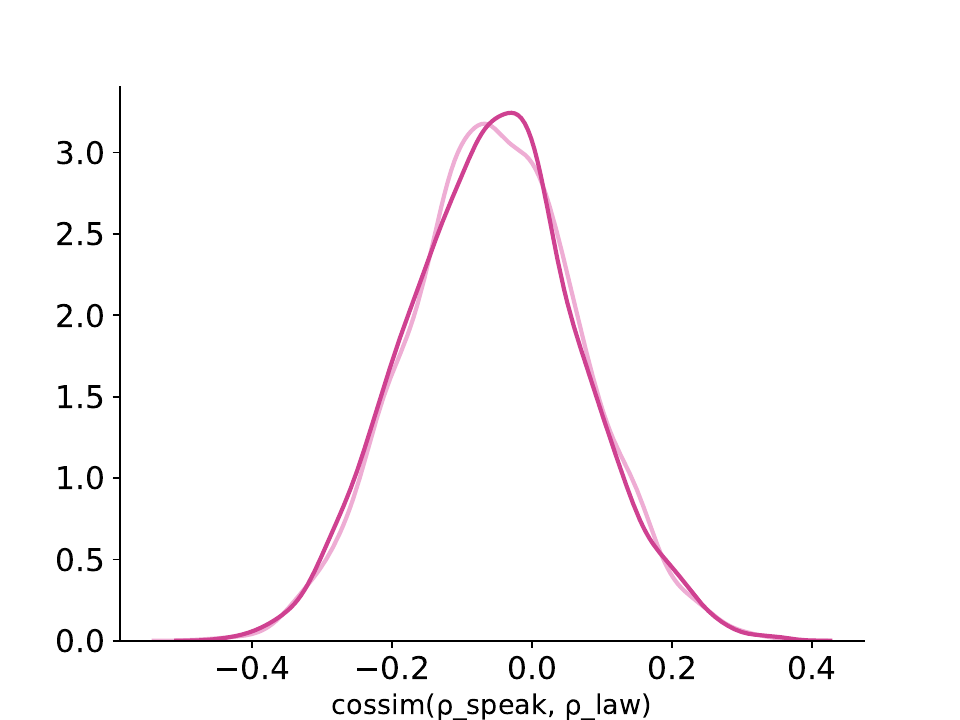} 
    \includegraphics[width=0.46\columnwidth]{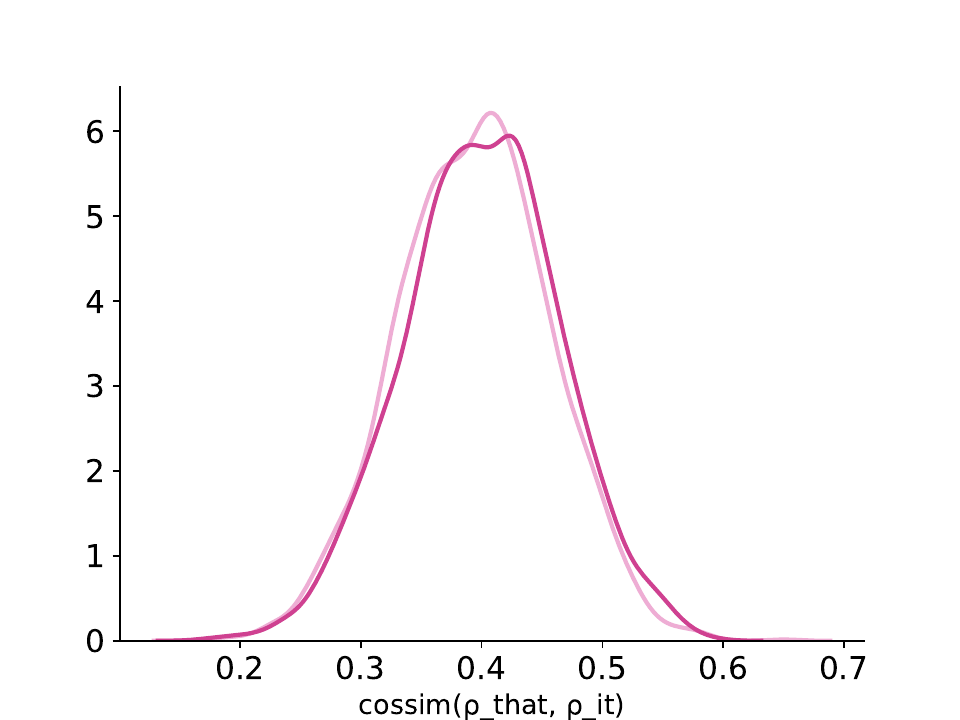} \\
    \caption{Posterior distribution of the cosine similarity between words. In dark, a Gibbs sampler with the last $K$ context vectors fixed, and in light, randomly selected $K$ words fixed to a MAP estimate. Left: 'speak' and 'law', right: 'that' and 'it'. $K=50, N=500,000, K=50$.}
    \label{fig:cossim_speak_law_sensitivity}
\end{center}
\end{figure}

To check the sensitivity of the results to the rotational fix, we calculated posteriors for the cosine similarity of some word vector pairs. In one run, we fixed the last $K$ context vectors as we do in the article. In another run, $K$ words were selected randomly from the vocabulary and their corresponding context vectors were fixed to the values of the MAP estimate. The results are shown in Figure \ref{fig:cossim_speak_law_sensitivity}. Visually, the resulting posteriors are very similar. Moreover, the $\hat R$ measures were very close to 1: $\hat R = 1.0012$ for both inspected word pairs, which is an indication of no difference between the two approaches.

\section{Laplace Approximation Implementation} \label{appendix:laplace_approx_details}

The Laplace approximation was implemented in the following steps
\begin{enumerate}
    \item Find the MAP estimate $\hat \theta_\textit{MAP}$
    \item Calculate the Hessian $\bm{H}(\theta)$
    \item Invert the Hessian $\bm{H}(\theta)$ at $\hat \theta_\textit{MAP}$ and set the covariance $\Sigma = \bm{H}(\hat \theta_\textit{MAP})^{-1}$
    \item Calculate the conditional covariance $\Sigma^\prime$ given that $\alpha_I = \hat \alpha_I$
    \item Find the eigendecomposition of $\Sigma^\prime = U DU^{-1}$
    \item Sample from $\theta \approx U \sqrt{D} \varepsilon$, where $\varepsilon$ is a vector of standard Gaussians. If some of the values of $D$ are negative due to numerical errors, set them to zero.
\end{enumerate}
The Cholesky decomposition failed due to the covariance not being numerically positive-definite. The eigendecomposition was used to find the closest numerically positive-definite decomposition $L^TL \approx \Sigma^\prime$. An alternative approach using pairwise covariances was also tested. After steps 1-4,
\begin{enumerate}
    \item For sampling $\rho_w$ and $\alpha_v$, formulate the covariance matrix $\Sigma_{\rho, \alpha} = \begin{bmatrix}
    \Sigma_{w,w} & \Sigma_{w,v} \\
    \Sigma_{v,w} & \Sigma_{v,v}
\end{bmatrix}$
    \item Sample $\begin{bmatrix}
        \rho_w \\ \alpha_v
    \end{bmatrix}$ from $\mathcal{N}(\begin{bmatrix}
        \hat \rho_w \\ \hat \alpha_v
    \end{bmatrix}, \Sigma_{\rho, \alpha})$ via the Cholesky decomposition
\end{enumerate}
The pairwise covariance matrices displayed less numerical issues, and the standard Cholesky decomposition could be used. The pairwise approach yielded similar results on the coverage experiment, albeit being slower.

\section{MFVI, MAP and HMC Implementation} \label{appendix:mfvi_and_hmc_implementation}

We implemented MFVI, MAP estimation, HMC using Stan. Specifically, we used \texttt{cmdstanpy} for MFVI and MAP, and \texttt{pystan} for HMC. See Listing~\ref{lst:unconstrained_stan} for the unconstrained model and Listing~\ref{lst:constrained_stan} for the constrained version, where we constrain the $K$ last context embeddings. Note that \texttt{lambda} used here denotes the standard deviation of the priors. We utilize the aggregated formulation of the likelihood, described by Equation \ref{eq:ll_aggregated}. 

For MFVI we use Stan’s default settings with 5000 iterations for optimization and draw 2000 posterior samples.
For MAP estimation, we use Stan’s optimization routine, with the default optimization algorithm L-BFGS.
For HMC we use 2 chains, each with 1000 burn-in and 2000 drawn samples.

All preprocessing and postprocessing was done in Python.

\onecolumn
\begin{lstlisting}[language=Stan, caption={Stan model for probabilistc skip-gram negative sampling}, label={lst:unconstrained_stan}]
data {
    real<lower=0.0> lambda; // prior standard dev.
    int<lower=1> U; // unique pairs count
    int V; // vocab size
    int K; // embedding dim
    array[U] int target_word;
    array[U] int context_word;
    array[U] int<lower=0, upper=1> posneg_labels; // 1: positive or 0: negative sample
    array[U] int counts; // counts of each unique pair
}

parameters {
    matrix[V,K] word_vectors;
    matrix[V,K] context_vectors;
}

model {
    // prior
    for (v in 1:V) {
        for (k in 1:K) {
            word_vectors[v, k] ~ normal(0, lambda);
            context_vectors[v, k] ~ normal(0, lambda);
        }
    } 

    // likelihood
    for (u in 1:U) {
        real l_contribution = dot_product(word_vectors[target_word[u]], 
            context_vectors[context_word[u]]);
        if (posneg_labels[u] == 1) {
            target += counts[u] * log_inv_logit(l_contribution);
        } else {
            target += counts[u] * log_inv_logit(-l_contribution);
        }
    }
}
\end{lstlisting}

\begin{lstlisting}[language=Stan, caption={Stan model for probabilistc skip-gram negative samples with $K$ constrained context embeddings}, label={lst:constrained_stan}]
data {
    real<lower=0.0> lambda; // prior standard dev.
    int<lower=1> U; // unique pairs count
    int<lower=1> V; // vocab size
    int<lower=1> K; // embedding dim
    array[U] int target_word;
    array[U] int context_word;
    array[U] int<lower=0, upper=1> posneg_labels; //  1: positive or 0: negative sample
    array[U] int counts; // counts for each unique pair
    
    // constraint K last context embeddings.
    matrix[K, K] fixed_context_matrix;
}

parameters {
    matrix[V, K] word_vectors;
    matrix[(V-K), K] context_vectors_raw;
}

transformed parameters {
    matrix[V, K] context_vectors;

    for (v in 1:(V-K)) {
        for (k in 1:K) {
            context_vectors[v, k] = context_vectors_raw[v, k];
        }
    }

    for (v in (V-K+1):V) {
        for (k in 1:K) {
            // row index in fixed_context_matrix is (v - (V-K))
            context_vectors[v, k] = fixed_context_matrix[v - (V - K), k];
        }
    }
}

model {

    // Only apply priors to the unconstrained the context vectors:
    for (v in 1:(V-K)) {
        for (k in 1:K) {
            context_vectors_raw[v, k] ~ normal(0, lambda);
        }
    }

    for (v in 1:V) {
        for (k in 1:K) {
            word_vectors[v, k] ~ normal(0, lambda);
        }
    }

    for (u in 1:U) {
        real l_contribution = dot_product(
            word_vectors[target_word[u]],
            context_vectors[context_word[u]]
        );
        if (posneg_labels[u] == 1) {
            target += counts[u] * log_inv_logit(l_contribution);
        } else {
            target += counts[u] * log_inv_logit(-l_contribution);
        }
    }
}

\end{lstlisting}

\section{Gibbs sampler Implementation} \label{appendix:mfvi_and_hmc_implementation}

The Gibbs sampler was implemented in Python using the TensorFlow framework and the \texttt{probabilistic\_word\_embeddings} PyPi package\footnote{(\url{https://pypi.org/project/probabilistic_word_embeddings/})}.

The Gibbs sampler was parallellized within iterations. Specifically, while sampling the word vectors $\rho_w$, 100 vectors were sampled at once. The matrix products, such as $\alpha^{\prime \top} \text{diag}(\omega)$, matrix inversions and other matrix and vector operations were all implemented in TensorFlow the 100 parallellized words at a time.
Moreover, sampling from the multivariate Gaussian was done via the Cholesky decomposition, using a parallellized implementation from TensorFlow
\begin{align}
    L_w L_w^T &= V_{\omega, w} \text{ (via Cholesky)}\\
    \beta_w &= L_w \varepsilon_w + \mu_\omega\\
    \varepsilon_w &\sim \mathcal{N}(0, \bm{I})
\end{align}

For the Polya-Gamma distribution, the CPU-based  \texttt{polyagamma} PyPi package\footnote{(\url{https://pypi.org/project/polyagamma/})} was used. Otherwise random variables were generated with TensorFlow.

\section{Computational Complexity} \label{appendix:computational_complexity}

The Gibbs sampler consists of six steps which are not negligible in terms of computational cost. The sampling cost per word can be expressed in terms of the dimensionality $K$, number of unique co-occuring words $n_u$, and the number of Polya-Gamma steps $S$.
\begin{enumerate}
    \item Initialization sampling -- $O(K)$
    \item Polya-Gamma random variable sampling -- $O(S \cdot n_u)$ (CPU)
    \item Calculating the precision matrix $V_\omega^{-1}$ -- $O(S \cdot n_u \cdot K^2)$
    \item Inverting the precision matrix to find $V_\omega$ -- $O(S \cdot K^3)$
    \item Cholesky decomposition $V_\omega$ -- $O(S \cdot K^3)$
    \item Sampling the vectors $\rho \sim \mathcal{N}(\mu, V_\omega)$ -- $O(S \cdot K^2)$
\end{enumerate}
To get the full complexity, most of these are just multiplied by the vocabulary size $V$. However, $n_u$ depends on the specific word. However, it occurs as a linear term, so $O(n_u \cdot V)$ evaluates to $O(N_u)$, where $N_u$ is the total number of unique word pairs in the data.

Thus, the total asymptotic complexity can be dominated by one of the following three factors
\begin{enumerate}
    \item Polya-Gamma random variable sampling -- $O(S \cdot N_u)$ (when the CPU dominates)
    \item Calculating the precision matrix $V_\omega^{-1}$ -- $O(S \cdot N_u \cdot K^2)$
    \item Inverting and decomposing the precision matrix -- $O(S \cdot V \cdot K^3)$
\end{enumerate}
In practice, the \texttt{polyagamma} package seems to be fast enough that the total complexity is
\begin{equation}
    O(S \cdot N_u \cdot K^2 + S \cdot V \cdot K^3)
\end{equation}
Note that by definition, $N_u \leq V^2$.

\end{document}